\documentclass[letterpaper]{article} 
\usepackage{aaai25}  
\usepackage{times}  
\usepackage{helvet}  
\usepackage{courier}  
\usepackage[hyphens]{url}  
\usepackage{graphicx} 
\usepackage{natbib}  
\usepackage{caption} 
\usepackage{algorithm}
\usepackage{algorithmic}
\usepackage{xcolor}
\usepackage{amsmath}  
\usepackage{amssymb}  
\usepackage{makecell}  
\usepackage{multirow}  
\usepackage{booktabs}
\usepackage{graphicx} 
\usepackage{caption}  

\usepackage{newfloat}
\usepackage{listings}
\DeclareCaptionStyle{ruled}{labelfont=normalfont,labelsep=colon,strut=off} 
\floatstyle{ruled}
\newfloat{listing}{tb}{lst}{}
\floatname{listing}{Listing}
\title{KPL: Training-Free Medical Knowledge Mining of Vision-Language Models}
\author{
    Jiaxiang Liu\textsuperscript{\rm 1, 2}, \ \ \  \ \ \
    Tianxiang Hu\textsuperscript{\rm 1, 2}, \ \ \   \ \ \
    Jiawei Du\textsuperscript{\rm 3, 4},   \\ 
    Ruiyuan Zhang\textsuperscript{\rm 1},  \ \ \ \ \ \
    Joey Tianyi Zhou\textsuperscript{\rm 3, 4}, \ \ \ \ \ \
    Zuozhu Liu \textsuperscript{\rm 1, 2}\equalcontrib 
}
\affiliations{
    \textsuperscript{\rm 1} ZJU-Angelalign R\&D Center for Intelligence Healthcare,  ZJU-UIUC Institute, Zhejiang University, China \\
    \textsuperscript{\rm 2} Zhejiang Key Laboratory of Medical Imaging Artificial Intelligence, Zhejiang University, China \\
    \textsuperscript{\rm 3} Centre for Frontier AI Research (CFAR), Agency for Science, Technology and Research (A*STAR), Singapore \\
    \textsuperscript{\rm 4} Institute of High Performance Computing (IHPC), Agency for Science, Technology and Research (A*STAR), Singapore

    \{jiaxiang.21, zuozhuliu\}@intl.zju.edu.cn
}

\usepackage{bibentry}

\begin{document}

\maketitle

\begin{abstract}
Visual Language Models such as CLIP excel in image recognition due to extensive image-text pre-training. However, applying the CLIP inference in zero-shot classification, particularly for medical image diagnosis, faces challenges due to: 1) \textit{the inadequacy of representing image classes solely with single category names}; 2) \textit{the modal gap between the visual and text spaces generated by CLIP encoders}. Despite attempts to enrich disease descriptions with large language models, the lack of class-specific knowledge often leads to poor performance. In addition, empirical evidence suggests that existing proxy learning methods for zero-shot image classification on natural image datasets exhibit instability when applied to medical datasets. 
To tackle these challenges, we introduce the Knowledge Proxy Learning (KPL) to mine knowledge from CLIP. 
KPL is designed to leverage CLIP's multimodal understandings for medical image classification through \textit{Text Proxy Optimization} and \textit{Multimodal Proxy Learning}.
Specifically, KPL retrieves image-relevant knowledge descriptions from the constructed knowledge-enhanced base to enrich semantic text proxies.
It then harnesses input images and these descriptions, encoded via CLIP, to stably generate multimodal proxies that boost the zero-shot classification performance.
Extensive experiments conducted on both medical and natural image datasets demonstrate that KPL enables effective zero-shot image classification, outperforming all baselines. These findings highlight the great potential in this paradigm of mining knowledge from CLIP for medical image classification and broader areas. 
\end{abstract}
\begin{links}
\link{Code}{https://github.com/JXLiu-AI/KPL}
\end{links}

\begin{figure}[h!]
\centering
\includegraphics[width=0.47\textwidth]{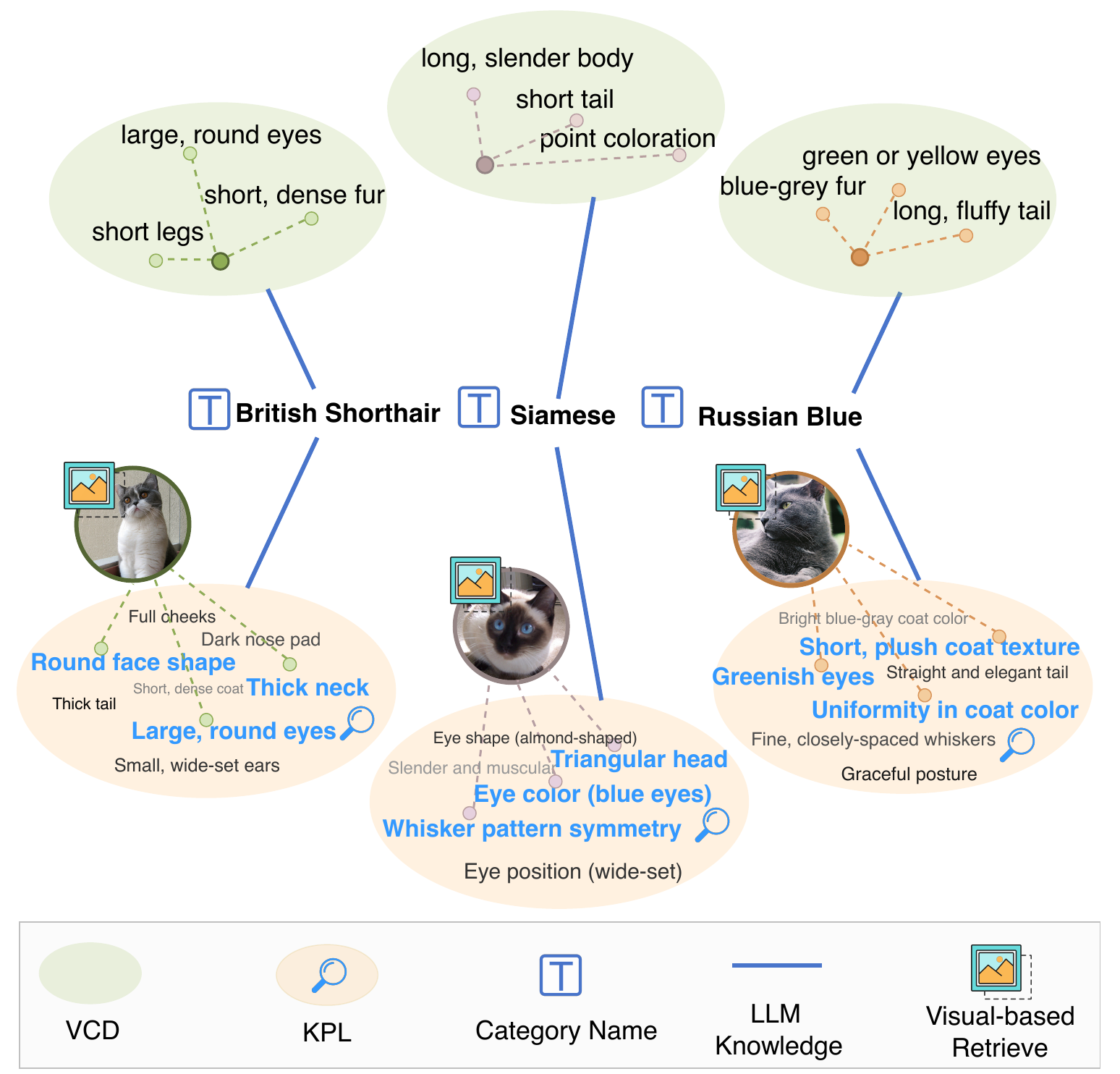}
\caption{VCD utilizes a limited number of descriptions that may be irrelevant to the images, whereas KPL generates richer descriptions from the knowledge of LLMs. These are visually-based and retrieved to ensure the descriptions' relevance to the images (Text Proxy Optimization).
} 
\label{Figure1}
\vspace{-1em}
\end{figure}

\section{Introduction}
Zero-shot image classification is a crucial task in various domains \cite{radford2021learning,wang2022clip,qian2023intra,ren2024chatgpt,zhou2023zegclip,jiao2023learning,novack2023chils,ali2023clip}, especially where comprehensive datasets covering all possible classes are unattainable and annotated images are scarce.
An exemplary application is zero-shot medical image diagnosis \cite{wang2022medclip,liu2023chatgpt,mahapatra2021medical,mahapatra2022self,xian2018feature,lampert2013attribute}, which is essential in real-world healthcare settings. 
Visual Language Models (VLMs) like CLIP have demonstrated promising performance in these tasks \cite{radford2021learning,liu2023parameter,eslami2021does,gai2024medthink,chen2024r,wang2023clipn}. After training CLIP on internet-sourced images with their corresponding captions or descriptions to gain image-text alignment \cite{radford2021learning,wang2023clipn,esmaeilpour2022zero}, the CLIP inference procedure utilizes class name features as proxies to classify images by comparing the similarity between image features and these text-based proxies \cite{radford2021learning}. 
While CLIP effectively uses unlabeled internet data for large-scale training to foster a rich multimodal understanding, its performance often falls short in specialized classification tasks, such as the medical domain \cite{liu2023chatgpt,lai2024carzero,liu2024vpl,liu2024medcot}.
This has led us to reconsider the existing CLIP inference paradigm.

We question whether the CLIP scheme fully utilizes its capabilities during inference. 
We contend that several facets of the vanilla CLIP inference methodology might substantially constrain the effectiveness of zero-shot image classification.
Firstly, using single class names as standalone representatives is questionable \cite{ren2024chatgpt,menon2022visual}, as CLIP models are trained to align images with descriptions, not just single class names (eg, synonyms might show up in the pre-training text). This simplification, while practical, is suboptimal and fails to fully leverage the rich semantic alignments developed during training.
Secondly, in zero-shot scenarios like medical image diagnosis, class names are often abstract, complex, and rarely present in CLIP’s pre-training data, posing challenges to the text encoder’s ability to interpret them accurately \cite{liu2023chatgpt}.
Thirdly, prior research indicates a persistent disparity between the text and vision feature spaces in CLIP \cite{qian2023intra,liang2022mind}, suggesting that direct similarity calculations may not fully capture the nuanced relationships needed for accurate classification.

\begin{figure}[t!]
\centering
\includegraphics[width=0.47\textwidth]{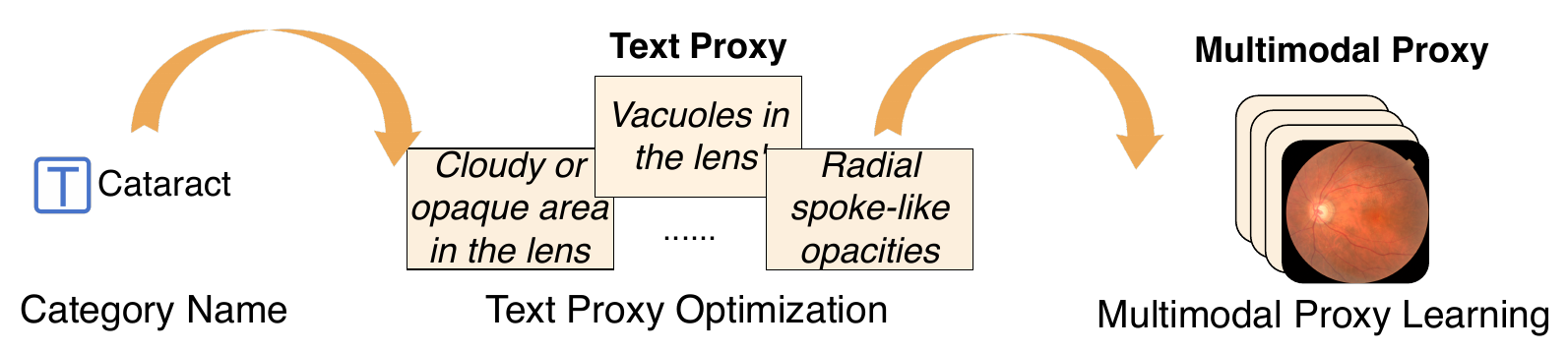}
\caption{
KPL is designed to leverage CLIP's capabilities for medical image classification through \textit{Text Proxy Optimization} and \textit{Multimodal Proxy Learning}.
First, semantic information is enriched for the cataract category through Text Proxy Optimization. Then, Multimodal Proxy Learning generates a multimodal proxy combining textual semantics and image knowledge to guide classification.
} 
\label{Figure2-2}
\vspace{-1em}
\end{figure}

Recent studies \cite{menon2022visual,ren2024chatgpt,liu2023chatgpt,novack2023chils,shiot} have been exploring how to leverage CLIP's knowledge. To address the inadequacy of using single class names for inference, Menon et al. and Ren et al. \cite{menon2022visual,ren2024chatgpt} proposed description-based classification methods that utilize large language models (LLMs) to enrich class names with descriptions to generate more sophisticated text proxies for conducting similarity calculations with images. In the medical field, Liu et al.\cite{liu2023chatgpt} enhanced CLIP models with disease details, mimicking human diagnostics. Despite the efforts of unleashing more of CLIP's textual ability via providing additional text, current description-based methods face the issue that the enriched descriptions are not semantically rich or relevant enough for the image classification tasks. For example, VCD \cite{menon2022visual} often misclassifies semantically similar categories, such as Siamese, British Shorthair, and Russian Blue in the Oxford Pets dataset \cite{zhang20220}. This misclassification stems from its reliance on descriptions like "short tail" and "long, slender body" which do not accurately reflect the images, compounded by the absence of relevant descriptors like "round face shape" and "thick neck". (see \textcolor{red}{Figure \ref{Figure1}}) These observations underscore the urgent need for developing a more precisely tailored description generation strategy.

To address the modal gap issue, Qian et al. \cite{qian2023intra} proposed learning proxies on the vision space, and employing the Sinkhorn to refine pseudo labels computed by text proxies, generated using the original category names, to guide the learning. However, current proxy learning methods, just like the CLIP inference, lacks text semantic richness, which can lead to inaccurate vision proxies that deviate from the intended optimal in the vision space. Moreover, as illustrated in \textcolor{red}{Figure \ref{pca}}, the Sinkhorn used for pseudo label refinement exhibits significant instability for medical image classification tasks, which highlights the need for a more robust algorithm in medical imaging to minimize information loss during the generation of pseudo labels which carry the knowledge from CLIP to guide the identification of optimal proxies across the modal gap.

In this paper, we propose a \textbf{K}nowledge \textbf{P}roxy \textbf{L}earning (KPL) to unlock CLIP's potential by mining knowledge from CLIP.
Specifically, KPL employs LLMs to build a knowledge-enhanced base for each category, ensuring semantically rich text resources.
Then, KPL enriches semantic information of category names through multi-level text and image retrieval for category-specific image-relevant knowledge from these comprehensive databases to optimize text proxies. Finally, The Stable GreenkHorn (SG) is proposed to refine the text proxies to guide the learning of multimodal proxies, which bridges the modal gap and results in improved decision boundaries. Comprehensive experimental results show that our method consistently outperforms all baseline methods on various datasets for diseases like pneumonia, glaucoma, cataracts, diabetic retinopathy, and malaria. 
Key contributions are summarized as follows:

\begin{itemize}
\item Our findings reveal a discernible gap between current methodologies and the theoretical optimal. This research advocates for a focused and in-depth exploration into the generation of multimodal proxy that exploits the existing capabilities of the CLIP model, with the ultimate goal of establishing a new CLIP-based inference paradigm. 

\item  We propose a two-step proxy optimization method, KPL.
KPL first builds semantic databases, optimizes text proxies with relevant image descriptions, and reduces semantic ambiguity between categories and images (Text Proxy Optimization). Afterwards, it optimizes multimodal proxies to address modal disparities and reduce performance degradation in medical contexts 
(Multimodal Proxy Learning).
\item  Extensive experiments and ablation studies conducted on nine datasets substantiate the efficacy of KPL. Notably, experiments reveal that KPL can also effectively mine knowledge from specialized domain CLIP such as BioMedCLIP \cite{zhang2023large} trained specifically on medical data to achieve performance improvements. These findings underscore the 
adaptability of the proposed paradigm.

\end{itemize}

\section{Preliminaries and Related Work}
\label{gen_inst}
\textbf{CLIP Inference}
Let $F, G$ denote the vision encoder and text encoder in CLIP~\cite{radford2021learning,wang2022medclip,zhang2023large,jeong2023winclip}. Denote the image dataset for inference as $\mathcal{D} = \{x_i\}_{i=1}^N$, and the category names as $\mathcal{C} = \{c_j\}_{j=1}^K$, the image and text features can be represented as $\{\mathbf{x}_i\}_{i=1}^N, \{\mathbf{z}_j\}_{j=1}^K$ where $\mathbf{x}_i = F(x_i), \mathbf{z}_j = G(c_j)$. The prediction of CLIP inference involves taking the features of category names as proxies $\mathbf{w}_j = \mathbf{z}_j, j = 1,...,K$ and calculating similarities between the image features and these proxies for classification $
\hat{y}_i = \arg\max_j (\mathbf{x}_i \cdot \mathbf{w}_j)$.

\textbf{Description-based Zero-shot Classification Method}
To leverage CLIP's comprehension of text, Menon et al. \cite{menon2022visual} and Ren et al. \cite{ren2024chatgpt} proposed to use LLM-generated category descriptions and a created multilevel knowledge tree as a supplement to category names. Liu et al. \cite{liu2023chatgpt} have explored description-based CLIP inference in the medical domain. These methods built enriched category text $\mathcal{C}' = \{ C_1, ..., C_K\}$ for classification based on category names $\mathcal{C} = \{ c_1, ..., c_K\}$, where each $C_j$ contains relevant text about $c_j$, and then took $\mathbf{w}_j = \frac{1}{|C_j|} \sum_{t \in C_j} G(t)$ as proxies.

\textbf{Proxy Learning}
Recent research \cite{qian2023intra,liang2022mind} have shown there is a gap between the text and vision space learned by CLIP. To reduce the modal gap, Qian et al. \cite{qian2023intra} introduced a proxy learning approach InMaP which obtains visual proxies from text proxies for classification via $\mathbf{w}_1, ..., \mathbf{w}_K = \underset{\mathbf{w}_1, ..., \mathbf{w}_K}{\mathrm{arg\,min}} \mathbb{E}_{x \sim \mathcal{D}} [KL(P'(x)||P(x; \mathbf{w}_1, ..., \mathbf{w}_K))]$, where $P(x; \mathbf{w}_1, ..., \mathbf{w}_K) = [\frac{\exp(\mathbf{x} \cdot \mathbf{w}_1 / \tau)}{\sum_{j=1}^K \exp(\mathbf{x} \cdot \mathbf{w}_K / \tau)}, ..., \frac{\exp(\mathbf{x} \cdot \mathbf{w}_K / \tau)}{\sum_{j=1}^K \exp(\mathbf{x} \cdot \mathbf{w}_K / \tau)}]$ denotes the class distribution calculated by the proxies to learn $\{\mathbf{w}_j\}_{j=1}^K$, and $P'(x)$ denotes the pseudo ground truth class distribution generated by text proxies $\{ \mathbf{z_j}\}_{j=1}^K$. Specifically, InMaP obtains the pseudo ground truth by solving the following optimization problem using Sinhorn algorithm \cite{sinkhorn1964,altschuler2017near}: $\max\limits_{P' \in \mathcal{P}}(\langle M, P' \rangle + \tau H(P')),$ $\mathcal{P} = \{P' | \forall i, \sum_j P'_{i,j} = \frac{1}{n}; \forall j, \sum_i P'_{i,j} = q_j; \forall i,j, P'_{i,j} \geq 0\},$ where $M_{i,j} = \mathbf{x}_i^T \mathbf{z}_j$, $H$ is entropy function of matrices, $q \in \mathbb{R}^K$ is a reference distribution over classes, and $\tau \in \mathbb{R}$ is the temperature.

\textbf{IPFP}
Iterative Proportional Fitting Procedure (IPFP), also known as biproportional fitting in various fields such as statistics, economics, and computer science \cite{stephan1942,RAS1965,idel2016review,chang2023}, has a rich history, being reexplored many times since its first appearing in \cite{yule1912,kruithof1937}. The attribution of the proof for uniqueness and convergence within this process is credited to Sinkhorn's seminal work \cite{sinkhorn1964}, which subsequently led to the naming of the algorithm in his honor. Recent efforts to analyze and enhance the Sinkhorn algorithm have led to developments like Greenkhorn \cite{altschuler2017near}. Studies \cite{lin2022efficiency,lin2019efficient,altschuler2017near} demonstrate that Greenkhorn retains similar theoretical guarantees and has better performance in practice.

\begin{figure*}[tb]
\centering
\includegraphics[width=0.85\textwidth]{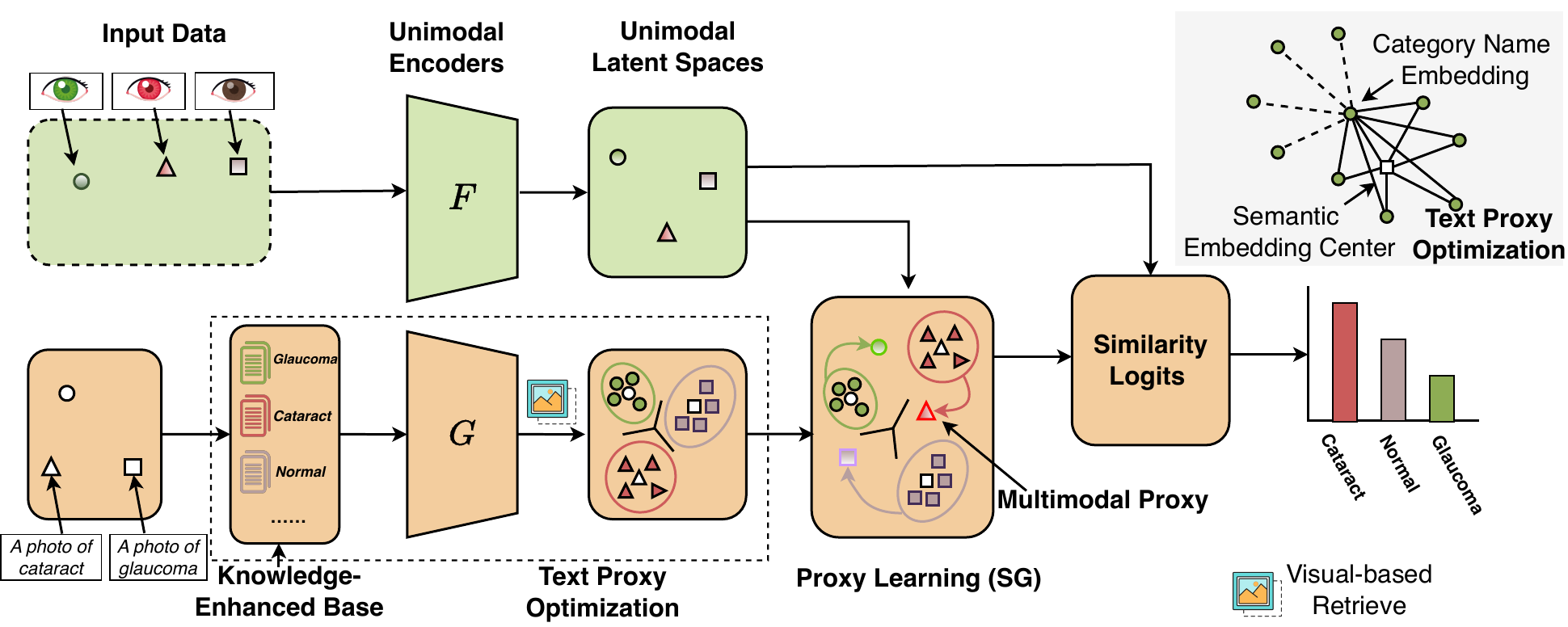}
\caption{Overview of KPL. 
1) Category texts (e.g., cataract) pass through a Knowledge-Enhanced Base and are encoded by a text encoder to produce semantic embedding centers via visual-based retrieval. Visual images are processed through a visual encoder to obtain visual features. 
2) CLIP uses the category name embeddings as text proxies to compare with visual features for classification results. 
3) KPL uses semantic embedding centers as text proxies to direct Multimodal Proxy Learning. Final classification results are obtained via the generated multimodal proxies and visual features. 
} 
\label{pipeline}
\vspace{-1.5em}
\end{figure*}

\section{Methodology}
\label{headings}

\subsection{Overview}
We focus on zero-shot image classification using CLIP~\cite{radford2021learning}. Recognizing the limitations of relying solely on category names and the existing text enrichment methods, as well as the issue of modality gaps, we propose a novel CLIP inference paradigm, KPL. Let $F, G$ denote the vision encoder and text encoder in CLIP~\cite{radford2021learning}. Denote the image dataset for inference as $\mathcal{D} = \{x_i\}_{i=1}^N$, and the class names as $\mathcal{C} = \{c_j\}_{j=1}^K$. The KPL pipeline, illustrated in \textcolor{red}{Figure \ref{pipeline}}, starts with $\mathcal{C}$, processed through a Text Proxy Optimization (TPO) which leverages CLIP capability, followed by encoding via text encoder $G$ to produce semantic embedding centers. Visual images $\mathcal{D}$ are processed by the visual encoder $F$ to obtain visual features. Subsequently, semantic embedding centers are employed to guide the Multimodal Proxy Learning to obtain multimodal proxies:


\begin{small}
\begin{gather}
    \mathbf{W}_{KPL}(\mathcal{D}, \mathcal{C}) = \underset{\mathbf{W}}{\mathrm{arg\,min}}\; \mathbb{E}_{x \sim \mathcal{D}} [d(Q_{F, G}(x; \mathcal{D}, \mathcal{C}), P_{F}(x, \mathbf{W}))], \\
    P_{F}(x, \mathbf{W})_{j} = \frac{\exp(F(x) \cdot \mathbf{w}_j / \tau)}{\sum_{k=1}^{K} \exp(F(x) \cdot \mathbf{w}_k / \tau)}, \forall j = 1,...,|\mathcal{C}| \nonumber
\label{eq1}
\end{gather}
\end{small}
$Q_{F,G}$ leverages CLIP and data to generate pseudo labels, and \(d\) is KL divergence by default.



Generated multimodal proxies $\mathbf{W}_{KPL}$ are integrated with visual features $\{ F(x)\}_{x \in \mathcal{D}}$ to produce classification results $\hat{y} = f(x) = \operatorname*{argmax}\limits_{j \in \{1,...,|\mathcal{C}|\}} F(x) \cdot \mathbf{w}_{KPL_j}, \forall x \in \mathcal{D}$. In the following sections, we introduce in details the \textit{Text Proxy Optimization} and the \textit{Multimodal Proxy Learning}.

\subsection{Text Proxy Optimization}
\label{KEB}

\textbf{Knowledge-Enhanced Base (KEB)}
To overcome the inadequacies of current text representations for categories (CLIP inference or current description-based CLIP inference), we propose to first utilize LLMs to generate category descriptions $\mathcal{C}_1 = \{C_1, \ldots, C_K\}$:  

\begin{equation}
C_j = \{t_{C_j}^1, \ldots, t_{C_j}^n \} = {\text{LLM}}(\text{prompt}, c_j),
\end{equation}
which build our knowledge-enhanced base.
These prompts should be designed to elicit a certain amount of category descriptions that aid in classification. In our experiments, the LLM used is ChatGPT, and the prompt for generating these descriptions is: \textit{According to published literature, what are useful medical visual features for distinguishing [Category] in an image? List as many features as possible, at least 50.} This approach aims to generate as many features as possible, while avoiding semantic loss due to the randomness inherent in LLMs.
For instance, for the category of [Cataract], we utilize LLMs to generate descriptive terms, such as ["Cloudy or opaque area in the lens"] and ["Vacuoles in the lens"], which we employ to construct a comprehensive knowledge base for [Cataract] (\textcolor{red}{Figure 2 and Appendix Figure 6}). 


\textbf{Visual-based Retrieval}
Despite having an extensive knowledge base, not all descriptions within it are directly related to images. Therefore, we further leverage CLIP's text-image alignment capabilities to retrieve descriptions that is more relevant to images for our classification tasks. Specifically, we select the top $k$ descriptions from the knowledge-enhanced base with the highest similarity to the image $x$, which finally forms $\mathcal{C}_2 = \{ C'_1, \ldots, C'_K\}$ ($\mathcal{C}_2 = \text{KEB}(\mathcal{C})$).

\begin{gather}
    C'_j = \{ t_{C_j}^{\sigma(1)}, \ldots, t_{C_j}^{\sigma(k)}\}, \\
    \mbox{where}\quad \{ s_{\sigma(1)}, \ldots, s_{\sigma(n)}\} = \operatorname{sort}_{\downarrow }\left( \{ s_1, \ldots, s_n \} \right), \nonumber \\
    s_l = \cos \langle \frac{1}{|\mathcal{D}|} \sum_{x \in \mathcal{D}} F(x), G(t_{C_j}^l) \rangle, \forall l=1,...,n. \nonumber   
\end{gather}
For example, for a cataract image, visual-based retrieval can identify from KEB relevant descriptions such as ["Cloudy or opaque area in the lens"] as shown in \textcolor{red}{Figure \ref{Figure2-2}}. 


As depicted in \textcolor{red}{Figure \ref{pipeline}}, the core of the Text Proxy Optimization involves constructing an extensive knowledge base (by setting the number requirement of descriptions in the prompts to LLMs) that encompasses detailed descriptions of each class, and leveraging CLIP’s existing alignment to identify the descriptions most helpful for classifying dataset $\mathcal{D}$. While $\frac{1}{|C'_j|} \sum_{t \in C'_j} G(t)$ can function as text proxies for classification by itself, the presence of a modal gap necessitates a more nuanced approach. Therefore, we propose to utilize it to guide the Multimodal Proxy Learning, leveraging it in a more effective manner.


\subsection{Multimodal Proxy Learning}
\label{proxylearning}

Following previous work on proxy learning, we use the solution of the optimization problem in the following to guide the learning of \textcolor{red}{Eq1}:

{\scriptsize
\begin{align}
    Q_{F, G}(\mathcal{D}, \mathcal{C}) &= \underset{P: P_{i,j} \geq 0, \sum_j P_{i,j} = \frac{1}{|\mathcal{D}|}, \sum_i P_{i,j} = q_j}{\mathrm{arg\,max}}\; \left(<M,P> + \tau H(P)\right), \nonumber
\end{align}
}

\begin{align}
    M_{i,j} &= F(x_i) \cdot \frac{1}{|C'_j|} \sum_{t \in C'_j} G(t), \quad C'_j \in \text{KEB}(\mathcal{C}), \nonumber \\
    &\quad \forall i = 1, \ldots, |\mathcal{D}|, \quad \forall j = 1, \ldots, |\mathcal{C}|
\label{eq4}
\end{align}

H denotes the entropy function, \( q \) in \(\mathbb{R}^{|\mathcal{C}|}\) is the reference distribution over classes, and \( \tau \) in \(\mathbb{R}\) is the temperature.

\textbf{Note} $Q_{F, G}(x; \mathcal{D}, \mathcal{C})$ indicates the row of $Q_{F, G}(\mathcal{D}, \mathcal{C})$ corresponding to image $x$.

\newtheorem{proposition}{Proposition}

\begin{proposition}
The optimization problem in \textcolor{red}{Eq4} has a unique solution of the form
\begin{equation}
P = diag(u) A diag(v),
\end{equation}
where $diag(u), diag(v)$ are two diagonal matrices with diagonals taken from vectors $u, v$, and $A = e^{\frac{M}{\tau}}$. See proof in Appendix.
\end{proposition}

Consider jointly the form of matrix $P$ from \textcolor{red}{Proposition 1} and the constraint on its rows and columns as delineated in the optimization problem, the optimization task essentially equates to estimate matrix $P$ given matrix $A = e^{\frac{M}{\tau}}$ such that
$
P_{i,j} = u_i A_{i,j} v_j, \ 
\sum_j P_{i,j} = \frac{1}{N} , \sum_i P_{i,j} = q_j.
$ This can be commonly achieved by the Sinkhorn algorithm ($\eta$ is iteration index of Sinkhorn) \cite{kruithof1937,sinkhorn1964}:

\begin{align}
\text{Set} \quad P^{(0)}_{i,j} = A_{i,j}, \text{and for all} \quad \eta \geq 1, \ \ \ \ \ \ \ \ \ \ \ \ \ \\
P^{(2 \eta -1)}_{i,j} = \frac{P^{(2 \eta -2)}_{i,j} * \frac{1}{N}}{\sum_{k=1}^K P^{(2 \eta -2)}_{i,k}}, P^{(2 \eta)}_{i,j} = \frac{P^{(2 \eta -1)}_{i,j} * q_j}{\sum_{k=1}^N P^{(2 \eta -1)}_{k,j}}.
\end{align} 

However, our empirical experiments indicate that pseudo labels generated by Sinkhorn did not consistently improve medical image classification performance, highlighting potential limitations in its efficacy (see \textcolor{red}{Appendix Figure 5}). Notably, prior research has explored the Greenkhorn algorithm \cite{altschuler2017near}, a greedy variant of Sinkhorn which maintains similar theoretical guarantees to Sinkhorn but has been shown to significantly outperform it in practice
\cite{lin2022efficiency,lin2019efficient,altschuler2017near}. Additionally, the traditional Sinkhorn algorithm faces computational challenges like underflow and overflow issues. Computing in logarithmic space can improve stability and increase error tolerance (see \textcolor{red}{Appendix Proof}), which is essential for handling the high dynamic range and potential extremes in medical imaging \cite{karim2004high,hung2013high}. Therefore, we proposed the SG algorithm for Multimodal Proxy Learning (\textcolor{red}{Appendix Algorithm 2}):

{\small
\begin{align}
\centering
\text{Set} \quad \log(P^{(0)}_{i,j}) &= \frac{M_{i,j}}{\tau}, \text{ and for all } 1 \leq \eta \leq N_{\text{max}}, \nonumber  
\end{align}
}

{\small
\begin{align}
\text{if} \quad &\underset{i}{\text{max}} |\sum_j P^{(\eta -1)}_{i,j} - \frac{1}{N}| > \underset{j}{\text{max}} |\sum_i P^{(\eta -1)}_{i,j} - q_j|: \nonumber \\
&r = \underset{i}{\mathrm{arg\,max}} |\sum_j P^{(\eta -1)}_{i,j} - \frac{1}{N}|, \nonumber \\
&\log(P^{(\eta)}_{r,j}) = \log(P^{(\eta -1)}_{r,j}) + \log\left(\frac{1}{N}\right) - \log\left(\sum_{k=1}^K P^{(\eta -1)}_{r,k}\right), \nonumber \\
\text{elif} \quad &\underset{i}{\text{max}} |\sum_j P^{(\eta -1)}_{i,j} - \frac{1}{N}| \leq \underset{j}{\text{max}} |\sum_i P^{(\eta -1)}_{i,j} - q_j|: \nonumber \\
&r = \underset{j}{\mathrm{arg\,max}} |\sum_i P^{(\eta -1)}_{i,j} - q_j|, \nonumber \\
&\log(P^{(\eta)}_{i,r}) = \log(P^{(\eta -1)}_{i,r}) + \log(q_r) - \log\left(\sum_{k=1}^N P^{(\eta -1)}_{k,r}\right). \nonumber
\end{align}
}

By selectively updating rows or columns with the greatest deviation from the target distribution, SG can address the inherent non-uniformity and complexity of medical datasets more effectively, thereby ensuring the preservation of knowledge mined from CLIP (see further analysis in \textcolor{red}{Appendix}). Consequently, the pseudo labels $Q_{F, G}(\mathcal{D}, \mathcal{C})$, which encapsulate this knowledge, guide the learning in \textcolor{red}{Eq1} to bridge the modal gap and establish multimodal proxies $\mathbf{W}_{KPL}$. Finally, the classification is conducted on the image dataset $\mathcal{D}$: $\hat{y} = f(x) = \operatorname*{argmax}\limits_{j \in \{1,...,K\}} F(x) \cdot \mathbf{w}_{KPL_j}, \forall x \in \mathcal{D}$.

\begin{table*}

\centering
\small
\vspace{-1em}

\begin{tabular}{ccccccc} 
\Xhline{1pt} 
\hline
                                &           & Shenzhen                                         & IDRiD                                        & MalariaCell                                  & Cataract                                      & Montgomery                                      \\ 
\hline
\multirow{5}{*}{ViT-L/14}       & CLIP \cite{radford2021learning}      & 50.76                                          & 06.80                                            & 48.53                                          & 16.64                                          & 60.14                                           \\
                                & VCD \cite{menon2022visual}       & 64.65                                          & 18.45                                          & 49.97                                          & 16.64                                          & 63.77                                           \\
                                & CMD \cite{liu2023chatgpt}       & 68.13                                          & 20.38                                          & 49.88                                          & 16.64                                          & 59.42                                           \\
                                & InMaP \cite{qian2023intra}     & 69.79                                          & 32.75                                          & 50.29                                          & 34.44                                          & 68.12                                           \\
                                & KPL (Ours) & \textbf{75.38} & \textbf{44.77} & \textbf{77.64} & \textbf{37.10}  & \textbf{71.74}  \\ 
\hline
\multirow{5}{*}{ViT-L@336px} & CLIP \cite{radford2021learning}      & 50.76                                          & 11.65                                          & 50.83                                          & 16.64                                          & 57.97                                           \\
                                & VCD \cite{menon2022visual}      & 69.94                                          & 18.45                                          & 50.00                                             & 16.80                                           & 62.32                                           \\
                                & CMD \cite{liu2023chatgpt}       & 68.88                                          & 13.59                                          & 49.92                                          & 16.80                                           & 57.97                                           \\
                                & InMaP \cite{qian2023intra}     & 73.72                                          & 38.37                                          & 54.66                                          & 31.11                                          & 60.87                                           \\
                                & KPL (Ours) & \textbf{78.55} & \textbf{47.09} & \textbf{80.86} & \textbf{44.43} & \textbf{70.29}  \\
\hline
\Xhline{1pt} 
\end{tabular}
\caption{Comparison of KPL with baselines across different backbones on medical image datasets.}
\label{maintable1}
\vspace{-1em}
\end{table*}

\begin{table}[h!]
\centering
\footnotesize
\resizebox{\linewidth}{!}{
\begin{tabular}{cccccc} 
\Xhline{1pt} 
                          &              & CUB                                               & Places365                                        & Oxford Pets                                     & ImageNet                                                 \\ 
\hline
\multirow{5}{*}{RN50}   & CLIP        & 49.02                                            & 32.48                                            & 74.60                                            & 54.85                                                    \\
                          & VCD          & 49.10                                            & 35.79                                            & 78.58                                            & 60.20                                                    \\
                          & Hierarchical & 49.86                                            & 36.83                                            & 79.99                                            & 60.63                                                    \\
                          & InMaP        & 47.38                                            & 40.21                                            & 86.08                                            & 63.17                                                    \\
                          & KPL$^{s}$(Ours)    & \textbf{50.50}    & \textbf{41.60}    & \textbf{88.81}   & \textbf{63.83}             \\ 
\hline
\multirow{5}{*}{ViT-B/16} & CLIP        & 56.66                                            & 38.21                                            & 81.68                                            & 63.53                                                    \\
                          & VCD         & 58.85                                            & 40.87                                            & 86.56                                            & 68.43                                                    \\
                          & Hierarchical & 59.25                                            & 41.52                                            & 87.19                                            & 68.95                                                    \\
                          & InMaP        & 57.31                                            & 43.79                                            & 91.79                                            & 71.92                                                    \\
                          & KPL$^{s}$(Ours)    & \textbf{59.49}   & \textbf{44.56}   & \textbf{92.67}   & \textbf{72.43}             \\ 
\hline
\multirow{5}{*}{ViT-L/14} & CLIP       & 63.46                                            & 38.65                                            & 87.92                                            & 76.60                                                    \\
                          & VCD         & 65.02                                            & 40.13                                            & 92.10                                            & 75.36                                                    \\
                          & Hierarchical  & 65.45                                            & 41.13                                            & 92.84                                            & \textbf{79.64}                                           \\
                          & InMaP        & 64.06                                            & 44.15                                            & 94.80                                            & 78.46                                                    \\
                          & KPL$^{s}$(Ours)    & \textbf{67.82} & \textbf{45.59} & \textbf{95.06} & 79.26                    \\
\Xhline{1pt} 
\end{tabular}
}
\caption{Comparison of KPL with baselines across different backbones on nature image datasets. KPL$^{s}$ denotes Multimodal Proxy Learning of KPL with Sinkhorn.}
\label{maintable2}
\vspace{-1em}
\end{table}

\section{Experiments}
\label{others}

\subsection{Experimental Setup}
\label{setup}
To validate KPL in medical zero-shot scenarios, we utilized several datasets including Shenzhen \cite{jaeger2014two}, IDRiD \cite{porwal2018indian}, MalariaCell \cite{hassan2022novel}, Cataract \cite{rokhana2022classification}, and Montgomery \cite{jaeger2014two}. These datasets cover a diverse range of diseases (\textcolor{red}{Appendix Table 1}), which align with previous research \cite{liu2023chatgpt}. 
Additionally, to test the versatility of KPL, we applied it to a wider array of natural image datasets including CUB \cite{wah2011caltech}, Places365 \cite{lopez2020semantic}, Oxford Pets \cite{zhang20220}, and ImageNet \cite{deng2009imagenet}. Classification accuracy is used as the evaluation metric in our experiments \cite{liu2023chatgpt,guo2024benchmarking}.

To ensure the applicability across different domains, we employed various pre-trained CLIP vision encoders for different datasets. For the medical image datasets, we used ViT-L/14 and ViT-L/14@336px. For the natural image datasets, we utilized RN-50, ViT-B/16, and ViT-L/14.
For the determination of key hyper-parameters, please refer to \textcolor{red}{Appendix Figure 2}. We construct the KEB using LLMs (GPT-3.5-turbo) to generate as many descriptions as possible for each category name, ensuring a minimum of \( n \geq 50 \) descriptions. 
The parameters \( k \), \( N_{\text{max}} \), and \( \tau \) are determined through Grid Search (\textcolor{red}{Apendix Figure 1}). Pseudocode is provided in \textcolor{red}{Algorithm 1 and 2}. 
We perform experiments in PyTorch framework on NVIDIA GEFORCE RTX 3090 GPU.

\begin{table}[t]
\centering
\resizebox{\linewidth}{!}{
\Large
\begin{tabular}{cccccccc} 
\toprule
KEB            & Retrieve                  & SG                 & Shenzhen       & IDRiD          & MalariaCell     & Cataract        & Montgomery      \\ 
\hline
                          & \checkmark & \checkmark & 76.13          & 39.15          & 31.00             & 31.78          & 68.12           \\
\checkmark & \checkmark &                           & 72.81          & 32.36          & 60.02          & 32.45          & 68.84           \\
                          &                           & \checkmark & 76.13          & 46.51          & 60.98          & 31.11          & 63.77           \\
\checkmark & \checkmark & \checkmark & \textbf{78.55} & \textbf{47.09} & \textbf{80.86} & \textbf{44.43} & \textbf{70.29}  \\
\bottomrule
\end{tabular}
}
\caption{Ablation study on different components with CLIP (ViT-L@336px) backbone on five datasets. Note that when operating without KEB and using visual-based retrieval with VCD descriptions \cite{menon2022visual}, \( k=5 \). Note that operating without SG denotes using KPL with Sinkhorn.}
\label{kb2}
\vspace{-1.5em}
\end{table}

\begin{figure}
\centering
\includegraphics[width=0.45\textwidth]{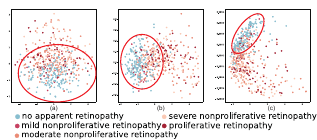}
\caption{PCA Visualization on the IDRiD with CLIP:
(a) CLIP using text proxies for feature visualization shows that the features of "no apparent retinopathy" are very dispersed and difficult to distinguish.
(b) Using Sinkhorn for Multimodal Proxy Learning, the features of "no apparent retinopathy" are slightly more clustered.
(c) Using the SG for Multimodal Proxy Learning, the features of "no apparent retinopathy" are tightly clustered into a distinct group.} 
\label{pca}
\vspace{-1em}
\end{figure}

\subsection{Main Results}
For medical datasets, we evaluate KPL and compare it to the state-of-the-art (SoTA) zero-shot image classification methods, including CLIP \cite{radford2021learning}, VCD \cite{menon2022visual}, CMD \cite{liu2023chatgpt}, and InMaP \cite{qian2023intra}, with two backbones on five datasets. 
Results are presented in \textcolor{red}{Table \ref{maintable1}}. We observe that our method consistently and significantly outperforms the baselines across all datasets and with any backbone used, demonstrating its effectiveness and robustness in medical image classification. This performance extends across various medical disease datasets, including tuberculosis, diabetic retinopathy, malaria cells, glaucoma, and cataracts. 
Notably, on the MalariaCell dataset, traditional methods struggled to accurately identify malaria cell images. However, KPL manages to effectively mine CLIP's knowledge to increase the accuracy by {50.83}\% to 80.86\%. This significant improvement transformed previously unrecognizable images into highly accurate identifications, suggesting KPL's potential as a new paradigm of employing VLMs for zero-shot medical image classification.

To validate the efficacy of the KPL paradigm, we compared KPL (with Sinkhorn) with SoTA methods such as CLIP \cite{radford2021learning}, VCD \cite{menon2022visual}, Hierarchical \cite{ren2024chatgpt}, and InMaP \cite{qian2023intra}, using three backbones on four natural image datasets, as shown in \textcolor{red}{Table \ref{maintable2}} (The reason for using KPL with Sinkhorn is explained in the \textcolor{red}{Appendix}). To ensure a fair comparison, we did not use any designed prompts as templates for the comparative methods.
We observed that in most cases, KPL outperforms the baselines across all datasets and backbones, demonstrating its high generalizability in various image domains. 

To demonstrate the improvements in representation learning brought by KPL, we perform PCA visualization \cite{abid2018exploring} according to the predictions of the model with CLIP (ViT-L/14@336px) on dataset IDRiD and compare the effects in \textcolor{red}{Figure \ref{pca}}. 
Subfigures (a), (b), (c) denote our models with different learning scheme versions, as described in the caption. 
We can observe from the visualization that KPL complete model (c) has distinct improvements over the model with naive learning schemes (b). 
As shown in \textcolor{red} {Figure \ref{pca} {(c)}}, the individual characteristics of the same type of disease are closer, and the feature distinction of different clusters of disease categories is more obvious, which shows the effectiveness of our designs for feature learning.

\subsection{Ablation Studies}

\textbf{Effects of KPL in Domain-Specific CLIP}
We applied KPL to BioMedCLIP (Note that only the BioMedCLIP (ViT-B/16) \cite{zhang2023large} model is publicly available), trained specifically for the medical domain, with results shown in \textcolor{red}{Table \ref{KPLMedical}}. 
Across all datasets and regardless of the domain knowledge used to train the CLIP \cite{radford2021learning} models, applying KPL with a training-free approach resulted in substantial improvements. This underscores the potential of KPL's inference paradigm to become a mainstream method for mining knowledge in future CLIP applications.
It is noteworthy that the CLIP (ViT-L/14), trained on internet data, surpassed the performance of BioMedCLIP (trained extensively on medical images) when enhanced with KPL. Furthermore, applying KPL to BioMedCLIP still significantly improves its classification performance.
Besides, as shown in \textcolor{red}{Figure \ref{vr}}, whether in medical datasets or natural image datasets, VCD experiences a performance leap after the application of Text Proxy Optimization, followed by another leap after Multimodal Proxy Learning. This shows that text proxy optimization and Multimodal Proxy Learning are both crucial steps in mining CLIP knowledge.

\begin{table*}
\centering
\small
\vspace{-1em}
\begin{tabular}{cccccc} 
\toprule
\textbf{Model}                                & Shenzhen                         & IDRiD                             & MalariaCell                       & Cataract                        & Montgomery                        \\ 
\hline
CLIP \cite{radford2021learning}                                          & 50.76                                     & 06.80                                     & 48.53                                     & 16.64                                     & 60.14                                     \\
 KPL (CLIP)       & 75.38\textcolor[rgb]{0,0.502,0}{(+24.62)} & 44.77\textcolor[rgb]{0,0.502,0}{(+37.97)} & 77.64\textcolor[rgb]{0,0.502,0}{(+29.11)} & 37.10\textcolor[rgb]{0,0.502,0}{(+20.46)} & 71.74\textcolor[rgb]{0,0.502,0}{(+11.6)}  \\
BioMedCLIP \cite{zhang2023large}                                    & 67.98                                     & 43.02                                     & 72.13                                     & 16.47                                     & 84.06                                     \\
 KPL (BioMedCLIP) & 77.79\textcolor[rgb]{0,0.502,0}{(+9.81)}  & 48.45\textcolor[rgb]{0,0.502,0}{(+5.43)}  & 91.61\textcolor[rgb]{0,0.502,0}{(+19.48)} & 51.25\textcolor[rgb]{0,0.502,0}{(+34.78)} & 89.86\textcolor[rgb]{0,0.502,0}{(+5.8)}   \\
\bottomrule
\end{tabular}
\caption{KPL on Medical-Domain CLIP.}
\label{KPLMedical}
\end{table*}

\begin{figure*}[h!]
\centering
\includegraphics[width=0.9\textwidth]{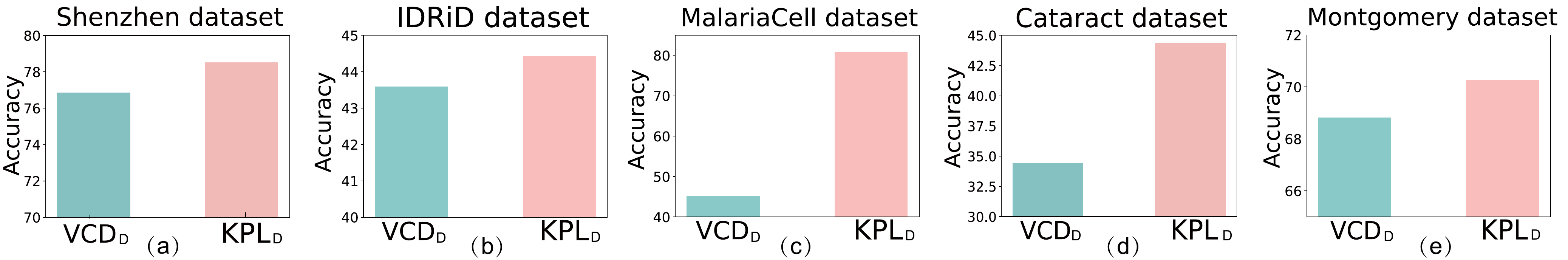}
\caption{Ablation studies with the KEB show that Multimodal Proxy Learning with KPL descriptions (KPL$_{D}$)  outperforms VCD (VCD$_{D}$), highlighting KEB's effectiveness.
} 
\label{kb1}
\vspace{-1em}
\end{figure*}

\begin{figure}[h!]
  \centering
  \includegraphics[width=0.42\textwidth]{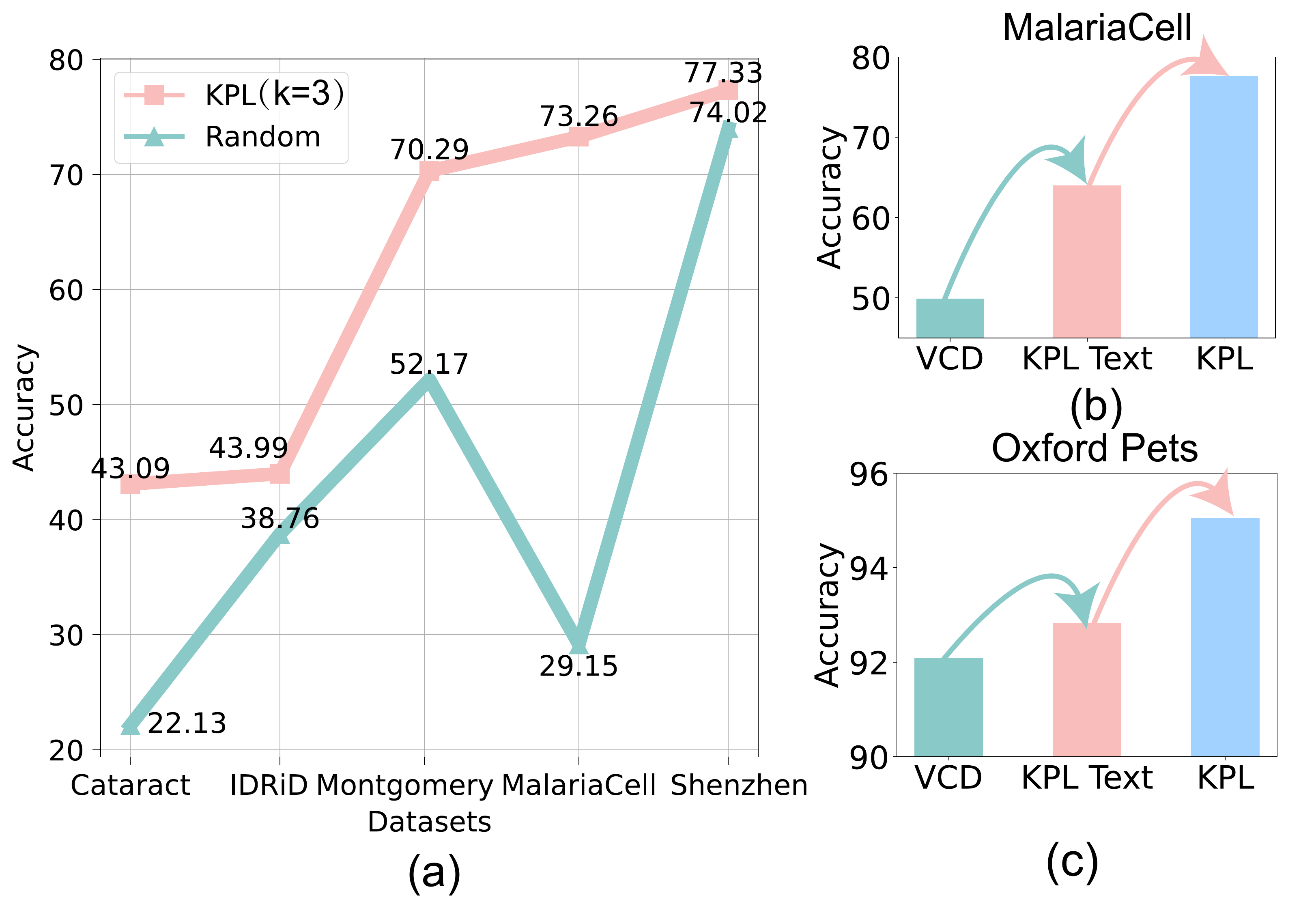}
  \caption{(a) Ablation study for visual-based retrieval at \( k=3 \) show that descriptions obtained through visual-based retrieval outperform randomly selected descriptions across five datasets.
(b),(c) On the MalariaCell and the Oxford Pets dataset, VCD performance significantly improves after Text Proxy Optimization (KPL Text) and again after Multimodal Proxy Learning (KPL).}

  \label{vr}
  \vspace{-1.5em}
\end{figure}

\textbf{Effects of KEB} 
We compared the performance of descriptions from VCD with those from the KEB, as shown in \textcolor{red}{Figure \ref{kb1}}. On five medical datasets, the best performance of visual-based retrieval enhanced by VCD descriptions still falls short compared to the performance of the KEB. 
This underscores the limitations of VCD's knowledge descriptions and highlights the superior guidance for classification learning provided by the KEB due to its richer knowledge content. 
\textcolor{red}{Table \ref{kb2}} also shows that KPL outperforms VCD in retrieval scenarios when \( k=5 \). Furthermore, in the \textcolor{red}{Appendix Figure 3}, grid search experiments for all \( k \) parameters in the Oxford Pets dataset between VCD and KPL further demonstrate the effectiveness of the KEB.

\textbf{Effects of Visual-based Retrieval} 
Within the KEB, we compared the experimental results of selecting retrieval descriptions with \( k=3 \) versus selecting three descriptions at random, as shown in \textcolor{red}{Figure \ref{vr}}. 
It is evident that on all medical imaging datasets, the performance of enhanced visual-based retrieval outperforms random selection.
Notably, on the MalariaCell, Montgomery, and Cataract datasets, the performance of randomly selected descriptions was very low. 
This could be due to the selection of descriptions irrelevant to the disease category in the image or even related to other diseases, leading to a significant drop in performance. This also demonstrates the effectiveness of visual-based retrieval. Additionally, in \textcolor{red}{Table \ref{kb2}}, results with $k = 5$ using only SG also surpass those of random description selection, underscoring the importance of visual-based retrieval descriptions. More detailed results regarding \( k \) can be found in the \textcolor{red}{Appendix Table 2}.

\textbf{Effects of SG} 
We also compared the traditional Sinkhorn \cite{sinkhorn1964} with the proposed SG, as shown in \textcolor{red}{Table \ref{kb2}} and \textcolor{red}{Appendix Table 3}. Replacing SG in KPL with the Sinkhorn resulted in performance declines across all datasets. The parameters for the Sinkhorn were set according to the reference \cite{qian2023intra}, demonstrating that stable updates to the most imbalanced rows and columns can preserve more knowledge and better guide the learning of multimodal proxies for medical image classification. Furthermore, we also conduct PCA Visualization \cite{abid2018exploring} to visualize features under three settings (original CLIP, Sinkhorn-guided proxy learning, and SG-guided proxy learning), as shown in \textcolor{red}{Figure \ref{pca}}. It can be observed that the PCA Visualization features of CLIP's text proxy remain dispersed and do not effectively distinguish between normal (blue) and diseased retinal features.
Using the Sinkhorn-guided multimodal proxy, features of the normal retina show some clustering. Finally, the SG-guided multimodal proxy allows the normal retinal features to cluster into a distinct group, as shown in \textcolor{red}{Figure \ref{pca} {(c)}}. 

\section{Conclusion}
In this work, we focus on mining CLIP's knowledge for zero-shot classification across multiple image domains. Our analysis reveals that the main obstacles to current classification performance are semantically insufficient text descriptions that are unrelated to the images and the modal disparity between visual and textual data. To address these challenges, we introduced a Text Proxy Optimization to generate more accurate descriptions, and then stabilize the pseudo label generation process by utilizing Multimodal Proxy Learning to reduce the modal gap more effectively. 
With KPL, the zero-shot accuracy of CLIP has been significantly enhanced on both medical image datasets and natural image datasets.

\section{Acknowledgments}
This work is supported by the Natural Science Foundation of Zhejiang Province, China (Grant No. LZ23F020008), the National Natural Science Foundation of China (Grant No. 62106222), and the Zhejiang University-Angelalign Inc. R\&D Center for Intelligent Healthcare.

\appendix




\bibliography{aaai25}

\end{document}